# An Intent-based Task-aware Shared Control Framework for Intuitive Hands Free Telemanipulation

Michael Bowman[1], Jiucai Zhang[2], and Xiaoli Zhang[1]*

**Abstract**
Shared control in teleoperation for providing robot assistance to accomplish object manipulation, called telemanipulation, is a new promising yet challenging problem. This has unique challenges—on top of teleoperation challenges in general—due to difficulties of physical discrepancy between human hands and robot hands as well as the fine motion constraints to constitute task success. We present an intuitive shared-control strategy where the focus is on generating robotic grasp poses which are better suited for human perception of successful teleoperated object manipulation and feeling of being in control of the robot, rather than developing objective stable grasp configurations for task success or following the human motion. The former is achieved by understanding human intent and autonomously taking over control on that inference. The latter is achieved by considering human inputs as hard motion constraints which the robot must abide. An arbitration of these two enables a trade-off for the subsequent robot motion to balance accomplishing the inferred task and motion constraints imposed by the operator. The arbitration framework adapts to the level of physical discrepancy between the human and different robot structures, enabling the assistance to indicate and appear to intuitively follow the user. To understand how users perceive good arbitration in object telemanipulation, we have conducted a user study with a hands-free telemanipulation setup to analyze the effect of factors including task predictability, perceived following, and user preference. The hands-free telemanipulation scene is chosen as the validation platform due to its more urgent need of intuitive robotics assistance for task success.

Introduction:

## A. Need for Robotic Assistance in Hands Free Teleoperation

Hands free teleoperation has seen recent attention as technological developments have surfaced. This form of teleoperation has been expanding due to the impact on broad applications such as telesurgery, assistive living, remote hazardous worksites, and distance learning. Common hands-free teleoperation approaches include using vision-based systems as the input, such as marker tracking or camera hand pose estimation. The cost of a workplace is dramatically reduced as RGB and RGB-D cameras are the only necessary hardware components (Li et al. (2019); Handa et al. (2020)). This provides a faster easier rollout and implementation. An example of hands-free teleoperation setup is shown in Figure 1a.

The difficulties in teleoperation stem from the indirect perception and indirect manipulation of the environment within the robot's workspace and the physical discrepancy between the operator and the robot (Rybarczyk et al. (2002); Healey (2008)). Telemanipulation of objects through controlling a robot hand are even more challenging because the additional difficulty of object interactions (Kumar et al. (2016); da Fonseca et al. (2019)). Most research efforts have focused

Michael Bowman[1] is a PhD Candidate in the Department of Mechanical Engineering at Colorado School of Mines, Golden, CO 80401 USA (email: mibowman@mines.edu).
Jiucai Zhang[2] is with the GAC R&D Center Silicon Valley, Sunnyvale, CA 94085 USA (e-mail: zhangjiucai@gmail.com)
*Xiaoli Zhang[1] is an Associate Professor in the Department of Mechanical Engineering at Colorado School of Mines, Golden, CO 80401 USA (*corresponding author, phone: 303-384-2343; fax: 303-273-3602; email: xlzhang@mines.edu).

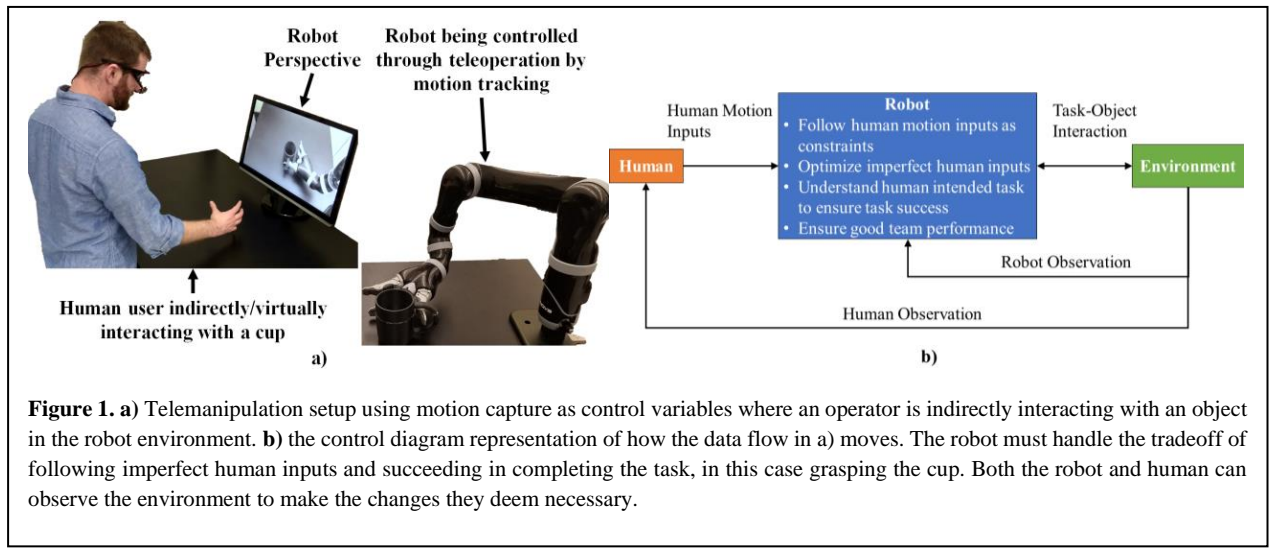

**Figure 1. a)** Telemanipulation setup using motion capture as control variables where an operator is indirectly interacting with an object in the robot environment. **b)** the control diagram representation of how the data flow in a) moves. The robot must handle the tradeoff of following imperfect human inputs and succeeding in completing the task, in this case grasping the cup. Both the robot and human can observe the environment to make the changes they deem necessary.

on the development of haptic feedback to survey the environment, accurate and unobtrusive human hand motion tracking, and virtual reality display. These approaches solved the indirect manipulation and indirect visualization issues of teleoperation by providing immersive, intuitive operations for the user such that the human better understands the remote environment and can provide cleaner inputs to the robot. In turn, allowing for better operation based on the human's sense of the environment.

However, hands free teleoperation pushes well established issues to the extreme. The indirect perception of the robot's environment is more challenging to overcome as common setups lack haptic feedback. This stresses the user to rely on visual feedback, and robot confirmations (Lin et al. (2020)). The indirect perception leads to the disembodiment problem of the user, reducing the effectiveness and reception of their robot partner. Physical discrepancy of the human hand and the robot hand also cause more difficulties in the control. There is no standard approach to defining the mapping between human and robot hands. This requires an operator to spend time to understand the behavior and feel comfortable using the system. Due to above reasons, the hands-free teleoperation is limited to relatively simple manipulation such as pick and place of simple objects, preset grasp routines initiated by user commands, fixed palm orientation control, and lack of in-hand manipulation.

### B. Lack of Shared Control in Telemanipulation

In the authors' vision, adding autonomous robotics assistance empowered by intent inference (so called shared control) is the most viable strategy for hands-free teleoperation to augment a person's operating ability to complete tasks as it uses indirect measurements from the human to inform the robot on actions it should undertake. Effectively, relying on the robot's own knowledge of the environment and allowing the robot to act in its own best interest to complete the operator's task as shown in the flow diagram of Figure 1b. Despite the advantages, intent inference for the purpose of grasping the target object has rarely been reported or investigated in telemanipulation. Another open problem of shared control for telemanipulation domains is to determine the tradeoff between task success (empowered by understanding human intent and providing intent-based robotic assistance) and following the operator (thus the operator feels in control of the robot). A tradeoff arbitration of these two is needed to enable the robot hand to flexibly reproduce actions that accommodate the operator's motion inputs as well as autonomously regulate the actions to compensate task constraints that facilitate subsequent manipulation.

### C.  Unique Challenge of Developing Shared Control for Telemanipulation

Telemanipulation has unique challenges as fine motion constraints dictate the task, intent inference, and legibility of the robot. Due to objects having multiple tasks for grasping, fine motion constraints are a way to decipher and disambiguate the difficult intent inference. However, this may not always be possible. For instance, a single grasp configuration can be used for multiple tasks such as relocation to another place or handing over to another person. The intent inference techniques—used in both data collection processes and controller design processes—in traditional shared control are less reliable as there is considerable overlap of the action space for telemanipulation. Without additional context or a way to consider the equivalency in task space, the redundancy in grasps can cause confusion for intent inference which disrupt shared control techniques traditionally used in teleoperation. Ideally, subtle fine motion constraints with differentiated intent can lead to multiple grasp solutions that equally solve the task. However, just because multiple grasps can solve the task, does not mean they are equally legible to an operator and bystanders. By choosing a grasp configuration that is less intuitive will result in adjustments and greater burdens on the operator. Thus, subtle motion constraints are an unique and instrumental factor to consider when inferring human manipulation intent and designing the arbitration policy of the shared control.

The subtle motion constraints are influenced by two factors: 1) the physical discrepancy of robot structures, and 2) the dependence of the robot specific grasp models. The physical differences of robot hand size, dexterity, and the number of fingers play a significant role in a person's perception of the robot's motion mimicking capability thus influencing the accuracy of intent inference as well as the perception of the arbitration strategy. Techniques and capabilities differ depending on the number of fingers, for instance, a 2 finger end-effector has less capability than a 5 finger anthropomorphic hand so the human operator expects less on the former to follow his/her motion. Further, the legibility needs to be more explicit with a hand that differs further from a human hand. This same principle extends to grasp models. Grasp models are inherently structure dependent where feasible actions should only be included. Having a grasp model with high amounts of varied actions allows for the most flexibility of applying subtle motion. These motion constraints bring extra difficulty to develop a generalizable shared control policy to handle the physical discrepancy of robot structures, and the dependence of the robot specific grasp models.

### D.  Evaluation of Human Perceived Robotic Assistance

Human perception of provided shared control assistance (general teleoperation, goal-guided approaches, or intent inference strategies) is critical to understand. Often, evaluating this assistance is done through task performance (i.e., improvement in task completion or time). Although task performance provides an objective means to evaluate the shared control assistance, it can only implicitly determine the human perception of the performance. Task performance is thought to be correlated to human perception where the faster a task is completed the more people like it. However, this correlation has been challenged repeatedly (Gopinath et al. (2017); Javdani et al. (2018); Young et al.(2019)). The legibility of the assisted actions, or feeling in control of the robot, is as critical if not more important in determining the effectiveness of provided assistance. Therefore, we identify an open problem to investigate the perception and legibility of assistance for two distinct groups, operators and bystanders. Ideally, operators will prefer a control mode that feels natural and intuitive, while being transparent to surrounding bystanders the intended actions the robot will take. *Understanding the legibility of both groups can help evaluate the tradeoff, between the task success and following the operator, of the shared control system.*

### E. Contributions

This paper focuses on developing a control scheme addressing the specific needs of shared control for object telemanipulation. This includes handling multiple tasks for the same object through intent inference and handling fine motion constraints from the operator inputs based on the robots' knowledge of the inferred tasks. The fine motion constraints also consider the physical discrepancy between the human and robot hands. The complexity difference between both hands is an important factor to consider as more complex structures can replicate actions of less complex structures due to having more degrees of freedom. Lastly, this paper investigates whether legibility for both operators and bystanders are equivalent and preferred. Although the scenario focused in this paper does not include haptic feedback devices, the control scheme presented in this work is not exclusively for hands free telemanipulation. We envision the autonomous assistance provided by our work can and should be paired with haptic devices for a complete shared control system.

The main contributions of this work are as follows:

1.) *Developed intent-based telemanipulation planner to handle multi-tasks for objects grasping.* We introduce an intent-based planner for the robot to generate grasp motion assistance toward achieving the inferred user's intent. In addition, the intent-aware planner handles ambiguous intent inference. The planner interprets an ambiguity level for the inference and considers it when generating a grasp configuration.
2.) *Generalized intent planners with constraints in an analogous approach for arbitration shared control.* By considering the domain of telemanipulation where human inputs (hand motion) are considered, we have created a general formulation which considers both achieving the intended task and following the operator's motion inputs. The formulation follows the intent-based planner and adds additional elastic constraints based on the importance of controllable variables.
3.) *Developed arbitration based on common features and physical discrepancy.* The physical discrepancy issue is necessary to address for different hand structures–which leads to different task constraints–in order to satisfy the goal. To combat the difference in the necessary task constraints between the operator and robot, we need to evaluate common features between their respective structures, thus generating an importance level for the robot to violate constraints to satisfy the goal. We present a method to quantitatively determine the difference between two separate hand structures. This method is used for selection or treated as a penalty for the constraints implemented. We have extended this to be generalized for different types of hand structures and robot end-effectors.

4.) *Investigated the perception and legibility of control from operators and bystanders.* Ranking questionnaire surveys of the operators (360 trials) and bystanders (1080 trials) gives insight on their view of favorability for the control framework. The operators need to feel in control of the robot for hands free teleoperation as there is a lack of physical perception of the robot environment. These feelings can be correlated with objective measures. The bystanders near an operational robot are an important aspect to consider. They may be weary of the actions by the robot and stop what they are doing to potentially assist the robot. This subjective analysis is important to consider as the goal of the telemanipulation is to improve the workflow instead of hindering it.

Related Work:

Shared control strategies have been utilized in reaching/approaching tasks (Michelman and Allen (2002); Kaupp et al. (2010);Mulling et al. (2015); Khoramshahi and Billard (2018);Losey et al. (2018)), yet have not been readily adapted to grasping and object manipulation domains. These approaches allow the operator to select the target to grasp. The robot chooses one of the predefined grasp types for the selected task. The methods to provide assistance in approaching–yet may not work as well in grasping scenarios–include envelope motion constraints (Abbott et al. (2007); Webb et al. (2016)), manually selective assistance levels (Feygin et al. (2002); Li and Okamura (2003)).

Intent-based shared control has shown promise and improvement in approaching tasks by placing the burden of task completion on the autonomous system (Losey et al. (2018)). The intent inference is measured through RGB (Song et al. (2013)), gaze (Li and Zhang (2017); Li et al. (2017)), and hand marker data (Li et al. (2020)). The measured intent inference is usually defined as a probability across a predefined set of tasks. The predefinition allows for easier classification and posterior probability assignment. To obtain the intent inference, independent data-driven classification models–Neural Networks (Burgard et al. (2015); Li et al. (2017, 2020)), Bayesian Networks (Hatakeya and Furuta (2003); Li and Zhang (2017); Li et al.(2020)), or Support Vector Machines (Park et al. (2016); Li and Zhang (2017); Li et al. (2020))– are developed from the total pool of human input data. Afterwards, shared control policies such as linear blending (Aarno et al. (2005); Dragan and Srinivasa (2013); Javdani et al. (2015)) are designed to arbitrate between robot intent inference and human motion. Linear blending strategies may not entirely work as the motion constraints from the operator's perspective and fully autonomous perspective may differ. Physical discrepancy between robot and human hands are normally ignored for the tele-approaching tasks.

These approaches normally employ a winner take all strategy based on the posterior probability for reaching tasks (Javdani et al. (2015); Reddy et al. (2018b); Losey et al.(2018)). These are justified to be used in reaching tasks as it normally pertains to disambiguate the inference between reaching different objects but may not be suitable to disambiguate the inference between different tasks of grasping the same object (grasp a cup for drinking or to transfer a cup to another person). Since most of these strategies employ winner take all for the highest probability or disambiguation of some form, they reduce the intent to a single task and do not handle solutions that solve multiple tasks. Lastly, the robot policy usually assumes more autonomy and control as the intent inference confidence improves (usually towards the end of a trajectory when the robot is near the goal). Since the robot gets added autonomy, the primary focus of the robot is to complete the task over listening to the operator's motion. These approaches can result in a perception of losing control of the robot as it may generate pre-defined grasp motion that appears different from human inputs towards the end of the grasp.

Pure intent-based formulations without shared control between human and the robot are inherently synonymous with autonomous control. The goal of these approaches is to use a robot's own understanding and grasping models to facilitate completing a human's intended target. These methods solely determine the best way to accomplish the task since they only are provided higher level information such as intended goal. In other words, it may be able to accomplish the same intended task yet accomplish it differently. This makes pure intent-based approaches ideal for implicit control techniques without feedback such as gaze control (Li and Zhang (2017); Li et al. (2017)) or gesture control (Li et al. (2019); Handa et al. (2020)) but may not be desired when a human operator provides explicit motion inputs like hand posture control.

Sematic grasping approaches inform a robot of functional task-specific grasps (Liu et al. (2020)). This allows a robot to understand a high-level task success before evaluating a lower level grasp success. A grasp can be considered successful based on criteria for grasp stability and contact pressure (Cutkosky (1989); Huebner (2012)), however, just because a grasp is successful does not mean it is appropriate. For instance, if one were to be at a tea party where it is socially acceptable to drink in a particular manner, others may be confused if you do not conform to their style. There are multiple ways to successfully grab a teacup to drink; however, it would result in a task failure if others do not perceive it as appropriate. Thus, one can think of task success as applying context for a grasp success (Song et al. (2010, 2015)). In (Song et al. (2011)), they demonstrate how to discretize and classify multiple classes with transitional grasp zones. The context aware grasping should be extended to what makes a grasp legible to their operator (i.e. grasping the tea cup in the appropriate manner). Within the same vein of task success for grasping, additional robot models (dependent on hand structures) are needed to successfully achieve the task in a way which their partner expects. The different grasp models generate different poses due to the physical discrepancy in end-effector design. For instance, the poses generated using the same intent inference for a two-finger robot and five-finger robot may be totally different from a human pose to achieve the same intent.

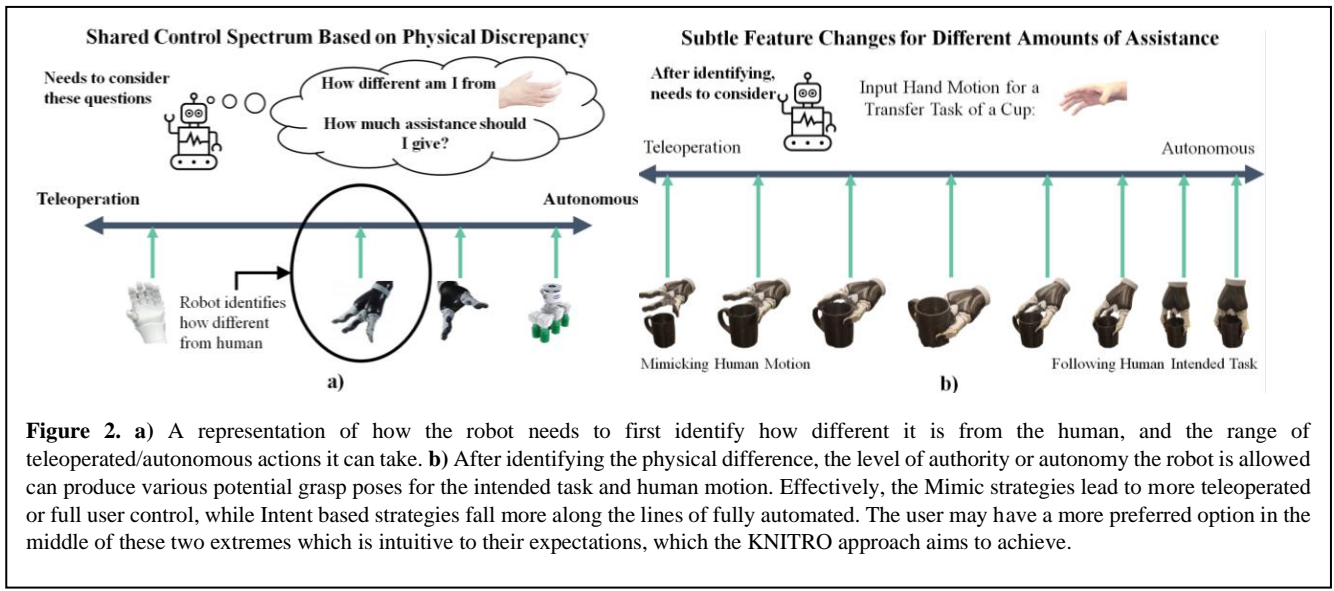

**Figure 2. a)** A representation of how the robot needs to first identify how different it is from the human, and the range of teleoperated/autonomous actions it can take. **b)** After identifying the physical difference, the level of authority or autonomy the robot is allowed can produce various potential grasp poses for the intended task and human motion. Effectively, the Mimic strategies lead to more teleoperated or full user control, while Intent based strategies fall more along the lines of fully automated. The user may have a more preferred option in the middle of these two extremes which is intuitive to their expectations, which the KNITRO approach aims to achieve.

Methods:

A. <u>Intent-based Shared Control Framework For Telemanipulation</u>

The main problem of shared control systems is determining the tradeoff or balancing between following the intuitive human motion and assisting in the task success. Shared control often is defined on a spectrum, where full user control and autonomous control are on opposite ends. A general sense of potential grasp configurations is demonstrated in Figure 2. Considering shared control as a spectrum from teleoperation (operator fully controlling the robot) to complete autonomy, we see different potential grasp configurations–some successfully do grasp the cup and others do not–which the operator may have a particular view on which is most successful. Through inferring the intended task, the autonomy can take over and focus on task success of the operator. This style of actions is best suited for control that lacks hand motion, i.e., eye gaze only control. Yet any of the poses presented in Figure 2 could be accepted for a particular operator and interface system. For instance, a well experienced operator may prefer minimal assistance with hand tracking, while a novice operator may rely on assistance as they get accustomed to the system.

To address the needs of the trade-off between balancing the task performance and following the operator, four separate components are developed and discussed in the following subsections. The four main components are: 1) disambiguation of the human intent for grasping tasks (section B) 2) building a robot model and formulating autonomous action for given intent inference (section B) 3) how to introduce motion constraints to generalize the intent planner (section C), and 4) how to determine arbitration between the motion constraints and intended task through common features and physical discrepancy (section D). The first two components formulate the intent module (top of Figure 3) which acts as an autonomous solution as seen in Figure 2. The latter two components deal with the motion constraint and determining the tradeoff.

B. Intent-based Telemanipulation Planner with Multi-Task Intent Inference and Grasp Modeling

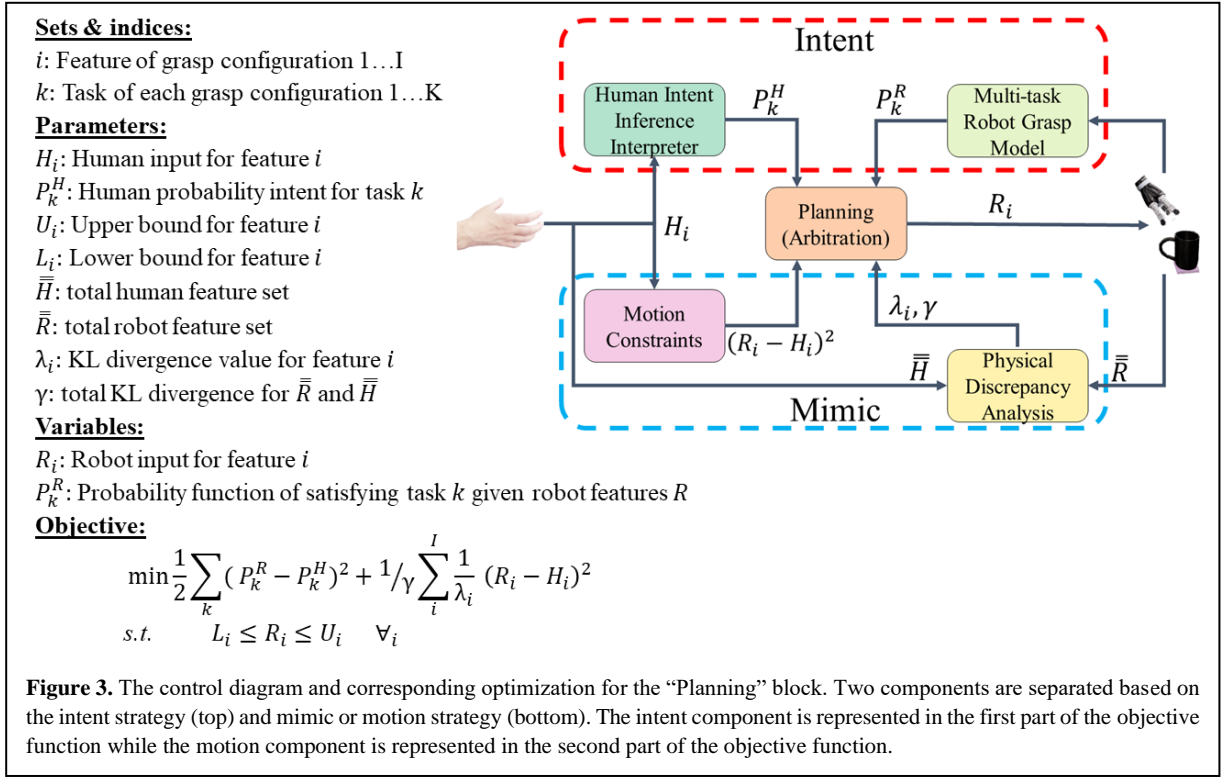

**Sets & indices:**
$i$: Feature of grasp configuration $1...I$
$k$: Task of each grasp configuration $1...K$
**Parameters:**
$H_i$: Human input for feature $i$
$P_k^H$: Human probability intent for task $k$
$U_i$: Upper bound for feature $i$
$L_i$: Lower bound for feature $i$
$\bar{\bar{H}}$: total human feature set
$\bar{\bar{R}}$: total robot feature set
$\lambda_i$: KL divergence value for feature $i$
$\gamma$: total KL divergence for $\bar{\bar{R}}$ and $\bar{\bar{H}}$
**Variables:**
$R_i$: Robot input for feature $i$
$P_k^R$: Probability function of satisfying task $k$ given robot features $R$
**Objective:**

$$\min \frac{1}{2}\sum_k (P_k^R - P_k^H)^2 + 1/\gamma \sum_i^I \frac{1}{\lambda_i}(R_i - H_i)^2$$

$$s.t. \quad L_i \leq R_i \leq U_i \quad \forall i$$

**Figure 3.** The control diagram and corresponding optimization for the "Planning" block. Two components are separated based on the intent strategy (top) and mimic or motion strategy (bottom). The intent component is represented in the first part of the objective function while the motion component is represented in the second part of the objective function.

In (Bowman et al. (2019)), we have validated that disambiguation of multi-inference tasks can be achieved through intent descriptor vectors. The following is how to use intent descriptors as a basis to formulate a shared control framework for hands free telemanipulation as shown in the top of Figure 3.

1. Intent Inference Interpreter and Descriptor Vectors

Consider the manipulation intent inference where there are three principle tasks, $w_m$, for grasping a cup where $w_m$, are three unique tasks: 1) using, or drinking from the cup, 2) transferring the cup to another location, and 3) handing the cup over to another agent. By stacking these probabilities together, a 3x1 vector is obtained where we can begin the development of a better descriptor vector. Since each classification model is obtained independently, it is not necessary for the total probability to equal one. For instance, in the remainder of the section assume the manipulation intent vector is [0.8,0.3,0.78]. In this instance, the $P(w_1)$ (the usage/drinking task) and $P(w_3)$ (the handover task) are almost identical to one another. There exist two forms of uncertainty, one from the human input, and another from the modeling process, thus it is critical to deal with this ambiguity by developing a descriptor vector. To establish the better descriptor or human target probability vector, $P_k^H$, by the following equation to combine the manipulation intent inference and reduce the ambiguity. Where $P(w)$ is the probability of the intent inference for the principle tasks

(i.e. drinking, transfer, and handover), and Y is a subset of the powerset of all combinations of the task, $\psi(m)$.

$$P_k^H = \prod_{i \in Y} P(w_i) \prod_{j \notin Y, j \in m} 1 - P(w_j), Y \subset \psi(m) \quad (1)$$

This results in $P_k^H$ being of size $2^m$ (in our example it is of size 8x1) because it accounts for overlap among multiple tasks. This equation describes each combination of the principle tasks as either true or false. From the example $w_m$ vector, the first case of the manipulation intent is "usage only". This would result in $P_1^H = 0.8(1-0.3)(1-0.78) = 0.1232$. Likewise, if the event were to satisfy all tasks, it would achieve $P_7^H = 0.8(0.3)(0.78) = 0.1872$. Figure 4 further demonstrates the relationships between the overlapping regions and how to calculate the probability. Upon calculating all possible combinations, the human target probability vector is treated as a reference the robot should attempt to match.

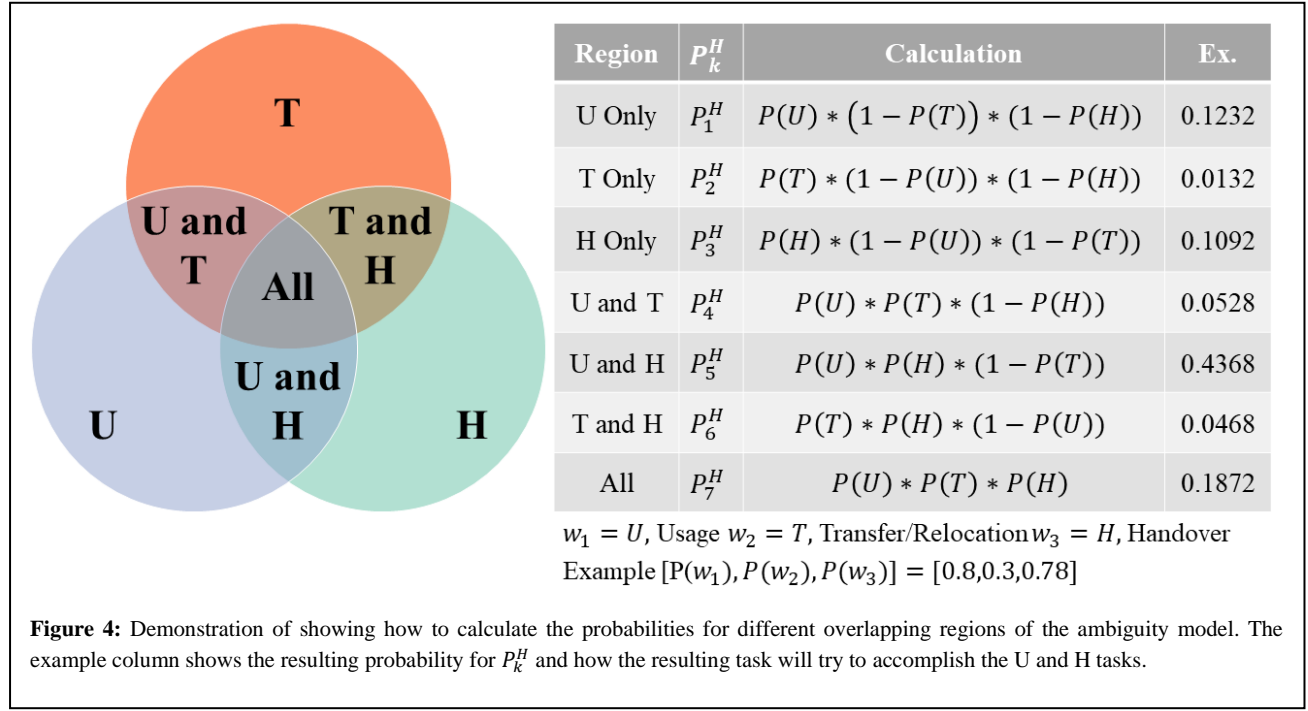

**Figure 4:** Demonstration of showing how to calculate the probabilities for different overlapping regions of the ambiguity model. The example column shows the resulting probability for $P_k^H$ and how the resulting task will try to accomplish the U and H tasks.

2. Multi-Task Robot Model

The robot planner also needs to produce its own probability vector of satisfying the intent inference distribution; where, for simplicity, it is based on the Naive Bayes robot model ($\bar{\bar{R}}$). Model $\bar{\bar{R}}$ is created by collecting feature vectors **R** for each task and using Expectation Maximization to determine their class distribution. Each vector **R** is comprised of individual features $R_i$ which contain robot pose (position and orientation), how much each finger should be open or closed (normalized between 0 and 1 where 0 is considered fully open). It can contain other features such as the robot joint angles, provided fingertip force and torque, as well as object features (shape, size, etc.). To produce the robot posterior probability vector, $P_k^R$, with a given **R**, the following two equations can be used where $\mu_k$ is the average value for task k, $\Sigma_k$ is the covariance matrix for task k. Both $\mu_k$ and $\Sigma_k$ rely on the dataset for $\bar{\bar{R}}$ and are obtained after the Expectation Maximization procedure. In equation 2, d refers to the dimension size of the feature vector **R**. Lastly, P(k) is determined by $\bar{\bar{R}}$, and the ratio of data points which the Expectation Maximization procedure assigned.

$$P(\mathbf{R}|k) = \frac{1}{\sqrt{\det(\Sigma_k) * (2\pi)^d}} e^{-\frac{1}{2}*(\mathbf{R}-\mu_k)^T * \Sigma_k^{-1} * (\mathbf{R}-\mu_k)} \quad (2)$$

$$P_k^R = P(k|\mathbf{R}) = \frac{P(\mathbf{R}|k) * P(k)}{\sum_k^K P(\mathbf{R}|k) * P(k)} \quad (3)$$

3. Formulation of Multi-Task Intent Controller.

By building the robot model with this soft assignment, the robot can understand grasping features within its own knowledge which can satisfy the different combinations of tasks. For instance, a single grasp can be used for both the drinking and handover of a cup. Further, we can take advantage of common grasp poses to satisfy the ambiguous manipulation intent ($w_m$). Upon developing the target probability vector and the robot probability vector, the objective function can minimize the difference. An L2 norm as shown in equation 4 is used to minimize the difference between the human target probability vector and robot probability vector. The objective is to have the robot match the human as closely as possible to ensure it completes the proper task.

$$min \frac{1}{2} \Sigma_k (P_k^R - P_k^H)^2 \quad (4)$$

Compared with the traditional framework for shared control involving approaching tasks, there is a need to handle solutions which are discontinuous, physically different from the human inputs, and come from complex interdependence models. The optimization proposed above handles discontinuous models through $P_k^R$ where multimodal distributions can be considered. Further, the model structure of $P_k^R$ can handle complex interdependence between inputs if the probability distributions are different from the Naïve Bayes model. Lastly, by using the intent disambiguation vectors in this encoding manner, the inputs from the human side ($H_i$) can differ from those of the robot side ($R_i$). However, moving forward in the next sections we will assume the type of inputs (robot pose or amount the fingers open and close) by the human and robot are the same.

C. <u>Generalized Intent Planner with Motion Constraints</u>

Two types of motion constraints must be considered. The first is the physical motion constraints of the robot where a workspace must be defined. The second are the motion constraints imposed by the human operator. The latter are more important for the determining the objective function of the shared control formulation which will be presented in the corresponding section.

1. Physical Motion Constraints of the Robot's Workspace.

The robot grasp model also contains upper and lower bounds for controllable variables, $R_i$ as show in equation 5. These are used for two reasons: 1) to bound the problem and help the planning process, and 2) to prevent external harm to the robot and the environment. For instance, to prevent external harm to the robot the table height can be turned into a lower bound, so the robot does not collide with it. The current formulation discussed only implicitly uses human motion to infer intent, however, it is necessary to consider explicit constraints to inform the robot on a particular manner to grasp an object.

$$L_i \leq R_i \leq U_i \quad \forall_i \quad (5)$$

2. Motion Constraints Imposed by the Human Operator.

An important factor to consider in the planning process is motion constraints an operator imposes. A simple type of motion constraint is to force all the operator's inputs to equal the controllable robot parameters as shown in equation 6. This is a strict case where an operator forces the robot to understand how a task should be accomplished—a necessary consideration to increase the effectiveness of assistance.

$$R_i = H_i \quad \forall_i \quad (6)$$

However, by explicitly dictating a constraint in this manner, it may lead to unintended errors as the burden of the control is placed on the operator. The operator could quickly become overwhelmed in hands free telemanipulation as they do not have sensory feedback and must determine how to map their own inputs to that of the robot's actions. This ultimately results in more readjustment to complete the tasks, or task failure. Further, defining motion constraints in this manner leads to the robot controller not providing effective assistance. The explicit nature does not allow the robot to utilize its own domain knowledge in exploring an alternative way to accomplish tasks. The motion constraints imposed could be outside of the bounds in the robot's grasp model knowledge leading to an infeasible solution, or an incorrect intent inference.

Alternatively, the robot should consider certain features to adhere to while also using its own knowledge to determine the best configuration to satisfy the task. By having an agent understand the relative differences between its own knowledge of control variables to another agent allows both systems to accommodate their own behaviors. This means, instead of forcing the operator to understand the robot's actions, we have the robot understand the human motion and pair this with the intent inference. Therefore, the robot is using its own knowledge to determine what are important constraints to follow and which are relatively less important. The hard motion constraints are formalized as elastic constraints, by combining equation 4 with an adjusted equation 6 to reformulate the objective function in equation 7.

$$\min \frac{1}{2}\Sigma_k (P_k^R - P_k^H)^2 + \frac{1}{\gamma}\Sigma_i^I \frac{1}{\lambda_i}(R_i - H_i)^2 \quad (7)$$

The benefits of this allow the robot to use its domain knowledge to alter how a user may want to accomplish a task. Further, the degree to which the robot violates the mimicking constraint is determined by a degree of commonality and importance between $H_i$ and $R_i$. The new components allow the robot to understand which features are common between itself and the human operator as well as how similar are these features. If both the human and robot features are near identical then the weights ($\gamma$, $\lambda_i$) approach 0 thus the overall weighted terms approach 1. When the penalty terms become dominant (approaches 1), the formulation begins to follow the mimic formulation where the user has full control over the robot. On the other hand, if the hands diverge and differ completely from one another, the weighted term goes to 0. This makes the intent matching aspect of the objective function dominant where it would resemble the intent formulation. Thus, with an objective function constructed in this manner, we have both components (autonomous actions and mimicking capabilities) needed for shared control along with an arbitration weight between the two.

### D. Arbitration Weighting Based on Common Features and Physical Discrepancy

Determining appropriate weights for common features is a two-fold process where not only common features are needed to be determined, but the degree in which the features are restricted. A way to analyze the similarity is to use Kullback-Liebler (KL) divergence (Hershey and Olsen (2007)). The goal of this divergence is to determine how two grasp models differ from one another. For this to work, assume the grasp model which is directly interacting with the object (the robot grasp model) is the true distribution, and the other (the human grasp model) is the inference to see how well one can predict the other. In other words, the human hand structure is used to infer the robot hand. However, the grasp models may diverge from one another differently. For instance, consider the robot grasp model ($\bar{\bar{R}}$) and human grasp model ($\bar{\bar{H}}$) where $\bar{\bar{R}}$ contains $\bar{\bar{H}}$ configurations in its own model ($\bar{\bar{R}} \supset \bar{\bar{H}}$). This would make predicting $\bar{\bar{R}}$ from $\bar{\bar{H}}$ easier than determining $\bar{\bar{H}}$ from $\bar{\bar{R}}$. Figure. 5 shows the general trend on how similarity using KL divergence works. Standard notation for KL divergence is KL (true population || inference population).

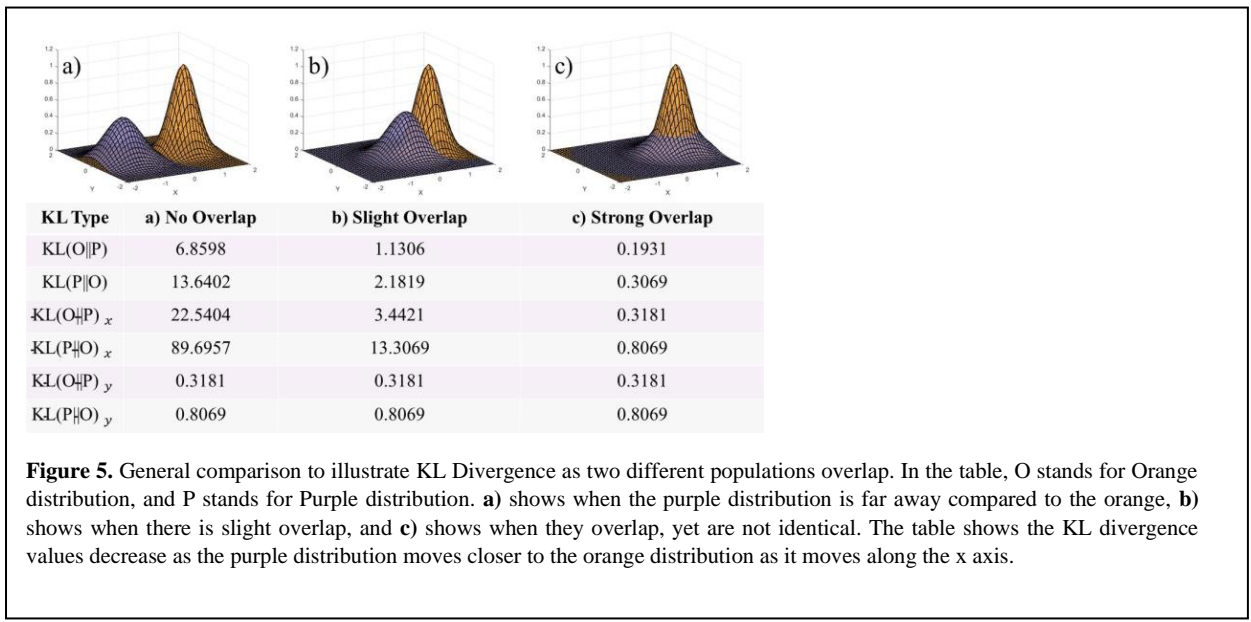

| KL Type | a) No Overlap | b) Slight Overlap | c) Strong Overlap |
|---|---|---|---|
| KL(O‖P) | 6.8598 | 1.1306 | 0.1931 |
| KL(P‖O) | 13.6402 | 2.1819 | 0.3069 |
| KL(O‖P) $_x$ | 22.5404 | 3.4421 | 0.3181 |
| KL(P‖O) $_x$ | 89.6957 | 13.3069 | 0.8069 |
| KL(O‖P) $_y$ | 0.3181 | 0.3181 | 0.3181 |
| KL(P‖O) $_y$ | 0.8069 | 0.8069 | 0.8069 |

**Figure 5.** General comparison to illustrate KL Divergence as two different populations overlap. In the table, O stands for Orange distribution, and P stands for Purple distribution. **a)** shows when the purple distribution is far away compared to the orange, **b)** shows when there is slight overlap, and **c)** shows when they overlap, yet are not identical. The table shows the KL divergence values decrease as the purple distribution moves closer to the orange distribution as it moves along the x axis.

In order to determine the commonalities between two populations of features, $\lambda_i$, and the entire hand configuration, $\gamma$, the KL divergence is used. Since all the grasp models created are assumed to be multivariate normal distributions, from the above sections, then each feature is univariate normally distributed. Thus, each population contains a mean, $\mu_{Hi}$ and $\mu_{Ri}$, and a standard deviation, $\sigma_{Hi}$ and $\sigma_{Ri}$. The KL divergence between the same feature from two separate populations (two different hand structures) can be defined as the following comparison between two univariate normal distributions.

$$\lambda_i = KL(H_i||R_i) = \ln\left(\frac{\sigma_{H_i}}{\sigma_{R_i}}\right) + \frac{\sigma_{R_i}^2 + (\mu_{R_i} - \mu_{H_i})^2}{2\sigma_{H_i}^2} - \frac{1}{2} \quad (8)$$

Where $\lambda_i$ are bounded from $[0, \infty]$ where 0 means there is no divergence, meaning the two populations are identical while infinity would mean they are completely different. This can be used to our advantage to determine the level of importance for each grasping feature between two separate hand configurations. Additionally, the multivariate normal distribution between two populations can be used to determine the overall divergence between hand configurations:

$$\gamma = KL(\boldsymbol{H}||\boldsymbol{R}) = \frac{1}{2}\left(Tr(\Sigma_H^{-1}\Sigma_R) + (\mu_H - \mu_R)^T \Sigma_H^{-1}(\mu_H - \mu_R) - d + \ln\frac{|\Sigma_H|}{|\Sigma_R|}\right) \quad (9)$$

Where d is the length of the feature vector **R** and where γ, is also bounded from $[0, \infty]$. Although these equations are specifically used for normally distributed variables, there are also known configurations for other distributions which may also prove to be effective for a more complex system.

Experimental Results:

A. Experimental Setup

We developed two controllers and use a third as a baseline. The first controller is only based on intent inference (equation 4) and does not include motion constraints from equation 6. This is referred to as the Intent Only. The second controller is based on the arbitration strategy in equation 7 which balances between the pure intent-based control and pure human mimicking control. This is referred to as Knowledge Intent Arbitration Optimization (KNITRO). The last is one is the baseline pure human mimicking controller which includes strict motion constraints of equation 6.

Two distinct groups are evaluated: the operators and the bystanders. The operators were asked to use a 3 finger MICO robot to accomplish 3 separate tasks with a cup, drinking or using the cup, the relocation or transfer of the cup, and handing the cup over to a bystander. An Xbox Kinect captures images of the operator's workspace (Figure 6), which were then fed into Google MediaPipe's (Lugaresi et al. (2019)) hand tracker to extract the hand. The hand is paired with pointcloud data from the Xbox Kinect to extract the real-world coordinates of the human hand. The output hand pose along with amount the fingers were open and closed were extracted as inputs for the controller optimization. The operators were told which task to perform, but they were free to grasp in their preferred way for each given task. The different preferred ways an operator may want to perform a task will give different intent inference. The operators were given training time to understand the three separate controllers. The operators were asked to complete six trials for each task for each controller. The total distance and time of the trajectories were recorded. They were also asked questionnaires after each trial. More information on the questionnaire is provided in the supplementary material. The grasps of the human and different controllers for the robot motion were recorded and saved for the bystanders to analyze later. This experiment was created to test three hypotheses:

1. Intent-based approaches (Intent Only and KNITRO) can outperform pure human control teleoperation in terms of success and completion time.
2. People prefer the arbitration approach (KNITRO) over the pure intent inference controller (Intent Only).
3. Human perception is going to be different for operators and bystanders.

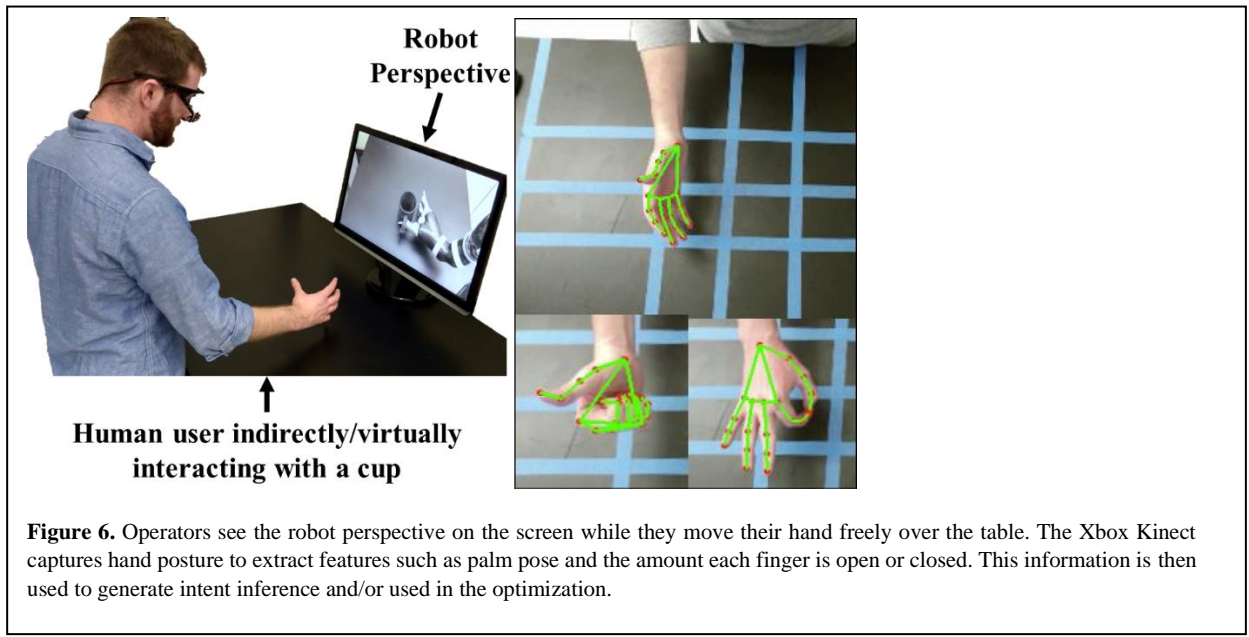

**Figure 6.** Operators see the robot perspective on the screen while they move their hand freely over the table. The Xbox Kinect captures hand posture to extract features such as palm pose and the amount each finger is open or closed. This information is then used to generate intent inference and/or used in the optimization.

### B. Operators analysis

Five operators completed the tasks in a predefined order of Mimic Only, KNITRO, Mimic Only, Intent Only. The Mimic Only approach acts as a baseline since it is pure telemanipulation, and it is critical for the operators to redo the baseline in between assistance modes as a reminder to the behavior of the baseline control mode. For the analysis, both Mimic Only tasks have been averaged together. Participants were allowed training time to feel comfortable with each control mode before the actual trials began. Participants were allowed to attempt to grasp the object as many times as possible until a successful grasp occurs or until the object fell over. In total, 360 trials were collected across the control modes. More information, including examples of successful grasp attempts for each controller, is shown in Extension 1 in the Appendix.

In summary of the following sections: The Intent and KNITRO approaches outperform the Mimic approaches in terms of success rate, completion time, and preference albeit task difficulty plays a factor in the significance of the impact. The KNITRO approach is the best for task success, while the Intent Only controller had the fastest completion times. The KNITRO strategy is slightly preferred to the Intent Only controller.

1. Success Rate

The KNITRO controller outperforms the other strategies. Due to the sample size, a Laplace estimate was used over Maximum Likelihood Estimate to better reflect the true success rate. The KNITRO controller has the best Laplace estimate (0.76), followed by the Intent strategy (0.63), and the Mimic controller was last (0.59). An adjusted-Wald 95% confidence interval was created for each controller was also calculated to see the expected bounds of success rate (shown in Figure 7). Lastly, a N-1 Chi-Squared test was conducted to determine if any of these controllers are greater than another and has statistical significance. No statistical significance was found between the Mimic and Intent strategies. The KNITRO approach has statistical significance with both the Mimic ($p=0.002$) and Intent ($p=0.026$) strategies. Adding assistance performs as well, if not better than traditional telemanipulation (Mimic strategy) in completing tasks.

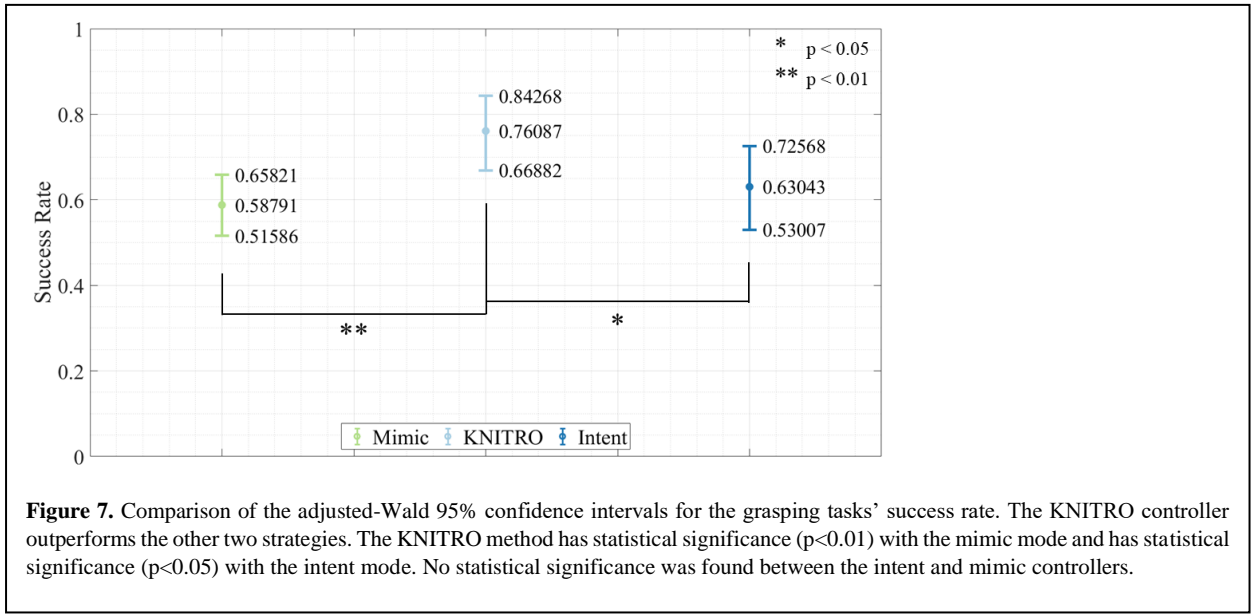

**Figure 7.** Comparison of the adjusted-Wald 95% confidence intervals for the grasping tasks' success rate. The KNITRO controller outperforms the other two strategies. The KNITRO method has statistical significance ($p<0.01$) with the mimic mode and has statistical significance ($p<0.05$) with the intent mode. No statistical significance was found between the intent and mimic controllers.

The summary of the task breakdown is presented in Table I and II. Table I shows the success rate, and confidence intervals for each task. In each instance, the KNITRO has the best the success rate. An important note is to remember the Mimic has more samples when developing the confidence interval which produces a much tighter interval. The assistance modes are at least competitive if not better than the Mimic approach.

Table I. Comparison of success rate and confidence intervals.

| Task | Controller | Laplace Success | CI Lower Bound | CI Upper Bound |
|---|---|---|---|---|
| All combined | Mimic | 0.588 | 0.516 | 0.658 |
|  | KNITRO | **0.761** | 0.669 | 0.843 |
|  | Intent | 0.630 | 0.530 | 0.726 |
| Transfer/Relocation  | Mimic   | 0.838 | 0.737 | 0.921 |
|                      | KNITRO  | **0.843** | 0.697 | 0.953 |
|                      | Intent  | 0.75  | 0.588 | 0.885 |
| Usage/Drinking       | Mimic   | 0.468 | 0.346 | 0.591 |
|                      | KNITRO  | **0.625** | 0.455 | 0.782 |
|                      | Intent  | 0.563 | 0.392 | 0.726 |
| Handover             | Mimic   | 0.452 | 0.331 | 0.575 |
|                      | KNITRO  | **0.781** | 0.623 | 0.909 |
|                      | Intent  | 0.563 | 0.392 | 0.726 |

With Table I, we see the separation of the success rate and confidence intervals grow as the tasks increase in difficulty (Transfer, Usage, Handover). Table II demonstrates the same sentiment through the statistical comparison across the tasks. For the first row, it seems much of the influence in differences across the controllers is based on the Handover task. The KNITRO outperforms the other two in difficult tasks, while easier tasks operators may not feel they need to rely on the assistance.

Table II. The statistical comparison of the task breakdowns for success rate

| Task Breakdown    | Intent vs. Mimic | Intent vs KNITRO | Mimic vs KNITRO |
|-------------------|------------------|------------------|-----------------|
| All combined      | 0.241            | **0.026**        | **0.002**       |
| Transfer/Relocate | 0.166            | 0.160            | 0.417           |
| Usage             | 0.187            | 0.300            | 0.069           |
| Handover          | 0.189            | **0.027**        | **0.0013**      |

2. Completion Time

The Intent and KNITRO controllers outperform the Mimic strategy. For this time analysis, only the successful trials were considered as failure times had high variance and would skew the data. Further, due to the nature of time data being positively skewed and not normally distributed (Sauro and Lewis (2010)), a log transform of the data is more appropriate. The geometric mean achieves the log-transformed times and it also tempers outliers better. The geometric means and confidence intervals are shown in Figure 8. A two-sample t-test was conducted to determine significance between control modes. No statistical significance was found between the Intent and KNITRO approaches. The Mimic approach had statistical significance with both the KNITRO strategy ($p = 0.0284$), and the Intent strategy ($p = 0.0002$). Providing autonomous assistance leads to the operators accomplishing the grasping tasks faster.

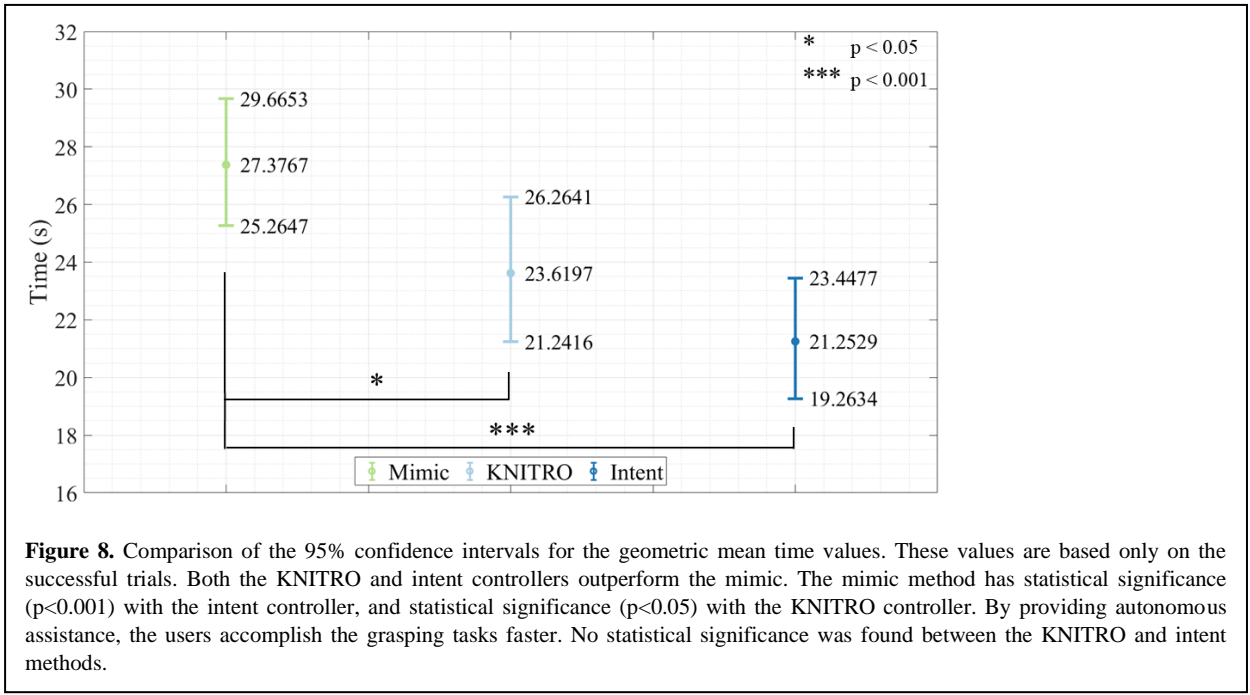

**Figure 8.** Comparison of the 95% confidence intervals for the geometric mean time values. These values are based only on the successful trials. Both the KNITRO and intent controllers outperform the mimic. The mimic method has statistical significance ($p<0.001$) with the intent controller, and statistical significance ($p<0.05$) with the KNITRO controller. By providing autonomous assistance, the users accomplish the grasping tasks faster. No statistical significance was found between the KNITRO and intent methods.

The summary of the task breakdowns is presented in Table III and IV. Table III shows the geometric means, and confidence intervals for each task. In each instance, the Intent has the best the completion times. The assistance modes outperform the Mimic approach in each task.

Table III. Comparison of success rate means and confidence intervals.

| Task | Controller | Geometric Mean (s) | CI Lower Bound (s) | CI Upper Bound (s) |
|---|---|---|---|---|
| All combined | Mimic | 27.34 | 25.26 | 29.67 |
| | KNITRO | 23.62 | 21.24 | 26.26 |
| | Intent | **21.25** | 19.26 | 23.45 |
| Transfer/Relocation | Mimic | 27.65 | 24.22 | 31.57 |
| | KNITRO | 26.66 | 22.42 | 31.70 |
| | Intent | **23.42** | 19.40 | 28.27 |
| Usage/Drinking | Mimic | 24.31 | 21.06 | 28.06 |
| | KNITRO | 20.34 | 16.78 | 24.67 |
| | Intent | **19.79** | 16.87 | 23.22 |
| Handover | Mimic | 30.39 | 26.69 | 34.59 |
| | KNITRO | 23.31 | 19.16 | 28.37 |
| | Intent | **20.01** | 17.00 | 23.55 |

With Table III, the assistance controllers separate from the Mimic controller as the task increase in difficulty. Table IV demonstrates the same sentiment through the statistical comparison across the tasks. For the first row results, the influence in differences across the controllers is based on the Handover task.

Table IV. The statistical comparison of the task breakdowns for completion task

| Task Breakdown | Intent vs. Mimic | Intent vs KNITRO | Mimic vs KNITRO |
| --- | --- | --- | --- |
| All combined | **0.0002** | 0.1452 | **0.0284** |
| Transfer/Relocate | 0.1460 | 0.3028 | 0.7306 |
| Usage | 0.0522 | 0.8150 | 0.1314 |
| Handover | **0.0002** | 0.2178 | **0.0252** |

3. Subjective analysis.

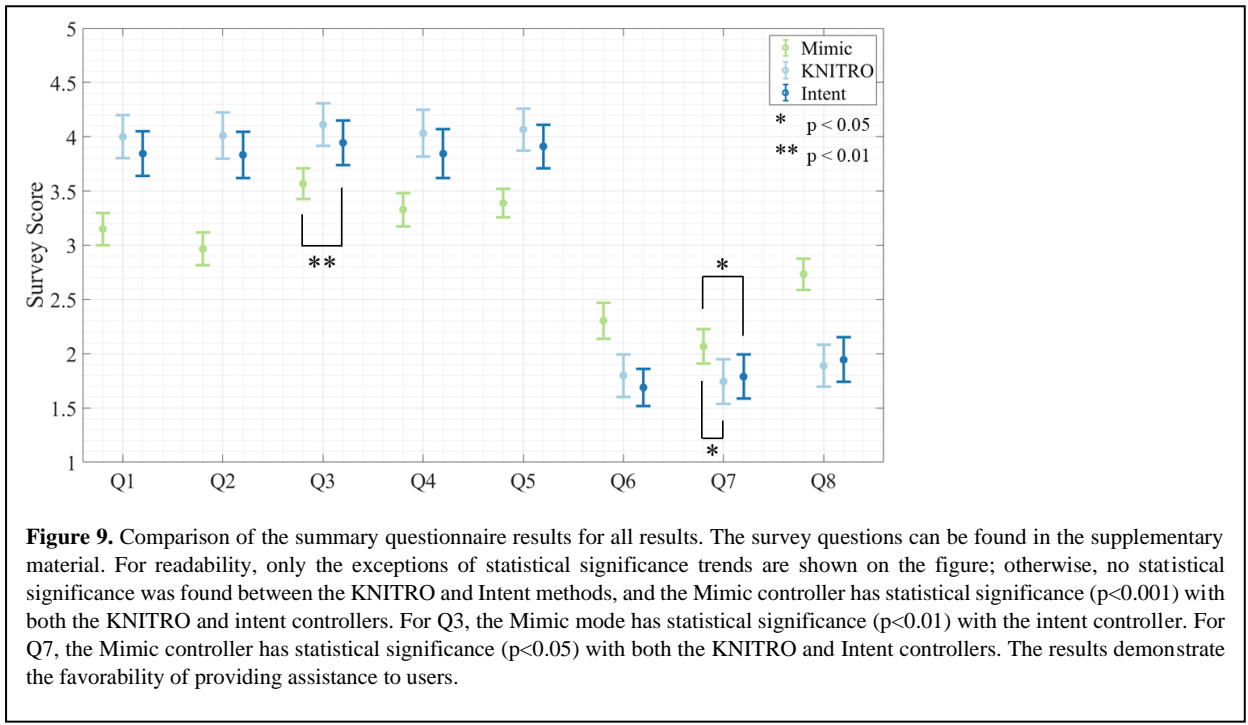

**Figure 9.** Comparison of the summary questionnaire results for all results. The survey questions can be found in the supplementary material. For readability, only the exceptions of statistical significance trends are shown on the figure; otherwise, no statistical significance was found between the KNITRO and Intent methods, and the Mimic controller has statistical significance ($p<0.001$) with both the KNITRO and intent controllers. For Q3, the Mimic mode has statistical significance ($p<0.01$) with the intent controller. For Q7, the Mimic controller has statistical significance ($p<0.05$) with both the KNITRO and Intent controllers. The results demonstrate the favorability of providing assistance to users.

The KNITRO and Intent approaches are more favorable than the Mimic controller. Operators were asked eight questions after each trial which they answer on a Likert scale from Strongly Disagree to Strongly Agree. The first five questions have positive connotation (higher score is better), while the last three have negative connotation (lower score better). The specific questions are displayed in the supplementary material. Figure 9 shows the t-distribution confidence intervals for each question. For readability, two trends have not been displayed. The first is the Mimic is statistically significant with the Intent strategy ($p=0.0001$), while the second is the Mimic strategy is statistically significant with the KNITRO strategy ($p = 0.0001$). The exceptions to these trends are in Q3 where the Mimic and Intent had statistical significance of $p = 0.0032$, and in Q7 where the Mimic strategy has statistical significance with the KNITRO strategy at $p=0.015$, and with the Intent strategy at $p=0.0344$. The assistance is preferred by the operators with no statistical difference between the two assistance strategies. For brevity, further task breakdown for the subjective analysis is displayed in the supplementary material.

C. Bystanders analysis

The 20 evaluators were asked to rank 54 different grasp configurations generated by the operators in the previous section. In total, 1080 trials across 20 evaluators were collected. The evaluators were not informed how any of the strategies behaved, nor were there any explicit identifying markers to determine the strategies. However, they were given a brief introduction to the abilities and limits of the MICO arm to ensure realistic expectations of the system. No evaluator was an operator of the robot. This is to ensure a fair comparison where the evaluators can see three potential grasp poses at one time instead of trying to recall each assistance strategy to compare against. The operators hand motion was displayed on a screen with the given target task next to the three final robot grasp configurations. The evaluators were asked to rank the three control strategies based on three separate criteria: 1) perceived task completion by the robot, 2) the perceived following of the operator, and 3) their overall preference.

1. Perceived Task Completion

The rank choices of the total task completion are show in Figure 10. The distributions of the rankings can inform some expected trends such as the Intent and KNITRO strategies are better at being transparent of the task completion. Based on the distributions presented in Figure 10, a Friedman test with a Least Significant Difference correction factor was determined to see statistical significance across the control strategies. The test was conducted with a standard p-value of 0.05. Table V shows the p-values of the Least Significant Difference statistical correction between the control strategies after the Friedman test found significance across the data. As expected, there is statistical significance between Intent and Mimic strategies which shows they are different. No statistical significance is observed with either strategy compared to the KNITRO.

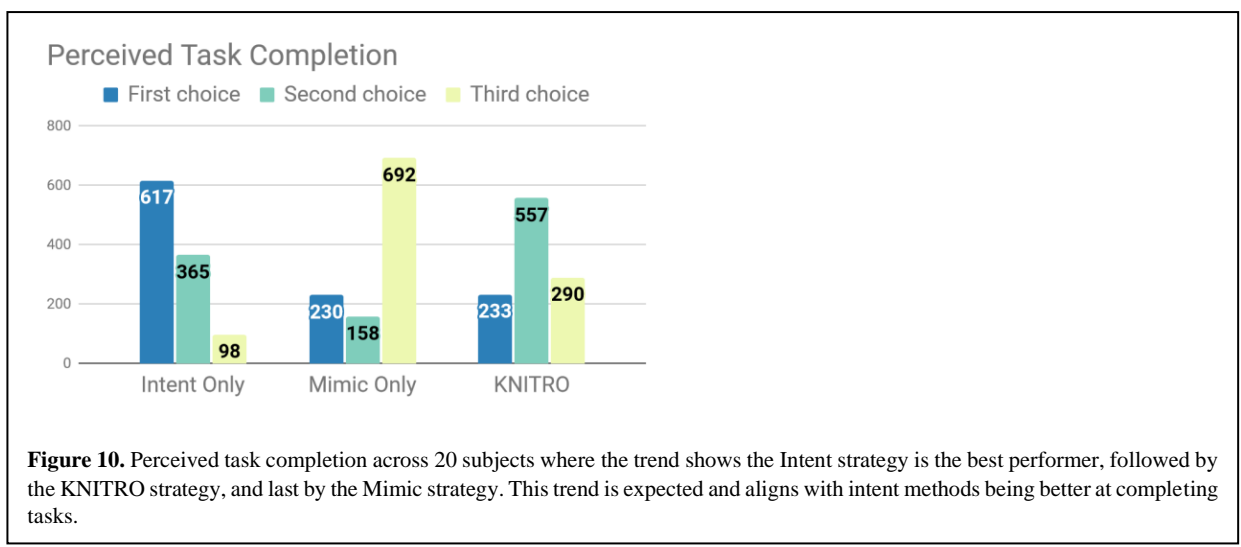

**Figure 10.** Perceived task completion across 20 subjects where the trend shows the Intent strategy is the best performer, followed by the KNITRO strategy, and last by the Mimic strategy. This trend is expected and aligns with intent methods being better at completing tasks.

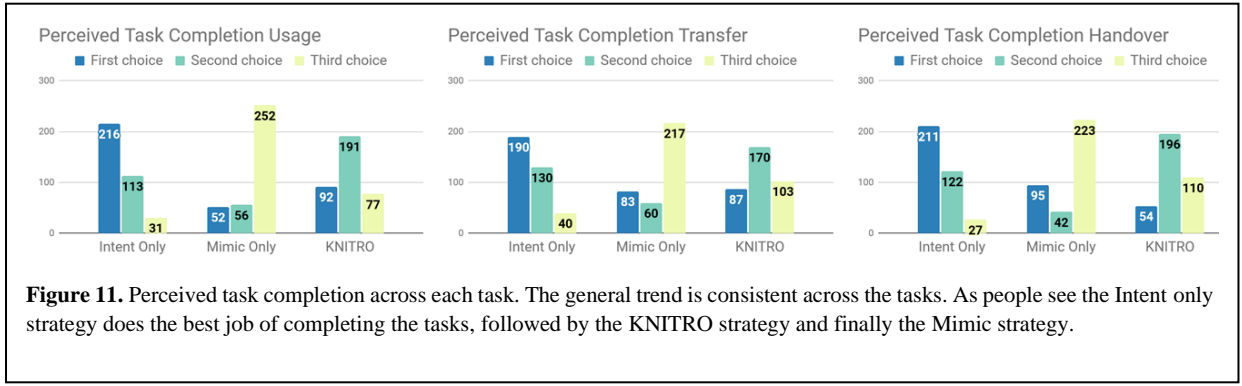

**Figure 11.** Perceived task completion across each task. The general trend is consistent across the tasks. As people see the Intent only strategy does the best job of completing the tasks, followed by the KNITRO strategy and finally the Mimic strategy.

Further, Figure 11 holds the individual task results for task completion. The Intent strategy is the perceived as the best across all tasks, while the Mimic strategy is perceived worst across tasks and the KNITRO is in between. Table V also shows the statistical significance across the individual tasks. The Intent and Mimic controllers continue to be statistically significant for each task. The individual datasets also show a greater separation between the Mimic and KNITRO approach for each task.

Table V: Statistical comparison across tasks for task completion

| Data Breakdown | Intent vs. Mimic | Intent vs KNITRO | Mimic vs KNITRO |
| --- | --- | --- | --- |
| Total | **4.53E-04** | **3.95E-02** | 1.48E-01 |
| Usage Only | **5.12E-04** | 2.45E-01 | **2.07E-02** |
| Handover Only | **4.89E-05** | 1.03E-01 | **1.51E-02** |
| Relocate Only | **8.03E-05** | 8.05E-02 | **2.81E-02** |

2. Perceived Following

The rank choices of the total perceived human following are shown in Figure 12. The trend is somewhat expected where the Intent only controller performs the worst of the strategies. It has an inverse relationship to the perceived task completion. This is expected as the Intent strategy does not follow the operator on how to accomplish the task and may choose a more well-suited manner to accomplish the task. The Mimic controller is the most favored. The higher third choice count is likely due to the evaluators feeling the mimicking grasp the robot was attempting was unnatural, and off putting due to finger placement. The KNITRO approach is competitive with the Mimic controller and is less susceptible to variational ranking since the grasp pose alters human input to better fit the robot grasp model. The similar first and second choices is likely due to the evaluators looking at more important features to accomplish the tasks such as the palm direction instead of smaller features such as finger placement. The KNITRO and the Intent strategy are near similar for the second choice. The KNITRO strategy sacrifices some of the ability to follow an operator in order to accommodate its own hand structure. Along with this, the Intent strategy implicitly follows the operator when the operator motion aligns with the robot model rules. No statistical significance was found across the following schemes.

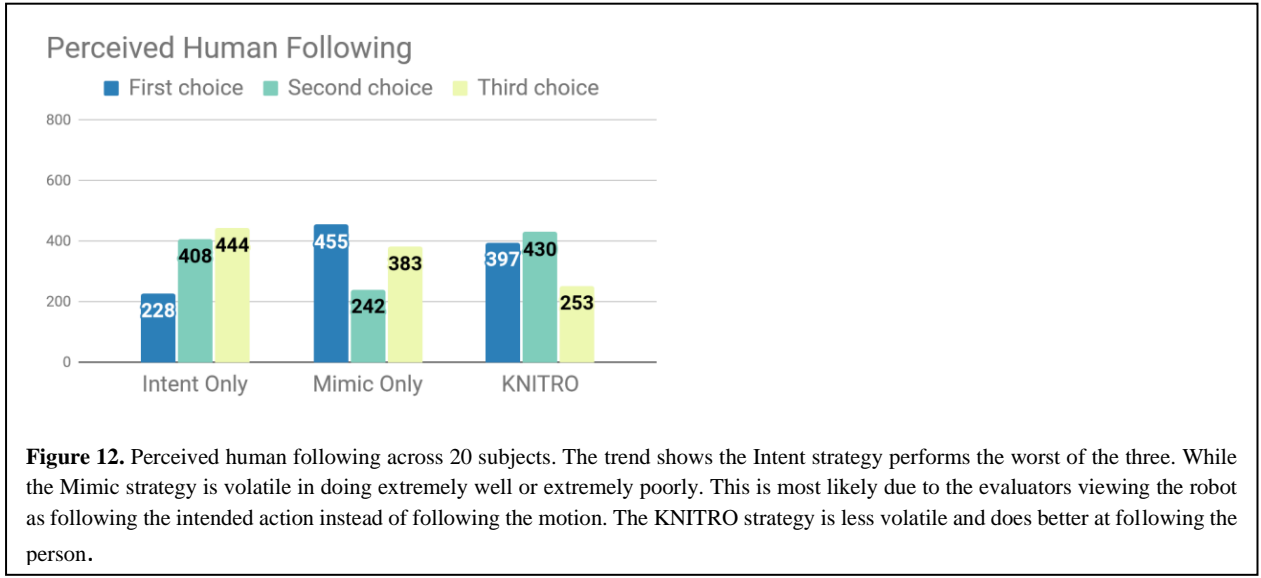

**Figure 12.** Perceived human following across 20 subjects. The trend shows the Intent strategy performs the worst of the three. While the Mimic strategy is volatile in doing extremely well or extremely poorly. This is most likely due to the evaluators viewing the robot as following the intended action instead of following the motion. The KNITRO strategy is less volatile and does better at following the person.

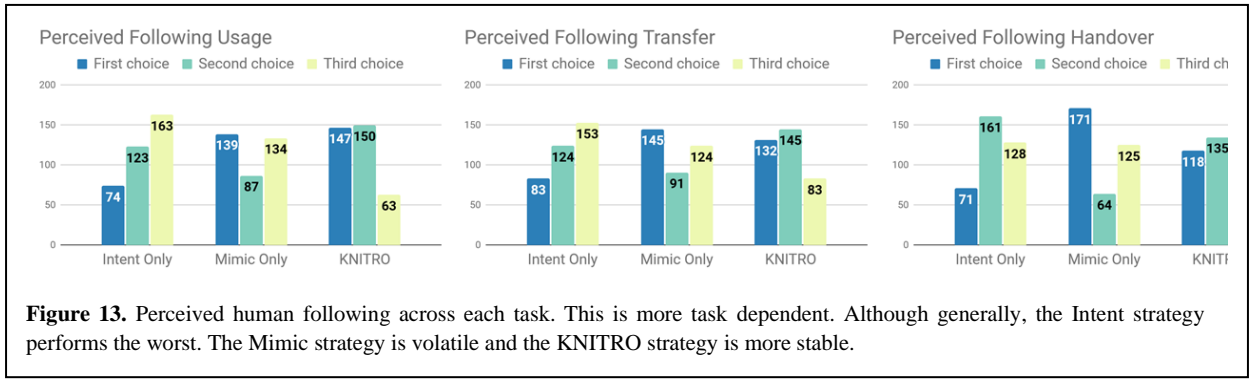

**Figure 13.** Perceived human following across each task. This is more task dependent. Although generally, the Intent strategy performs the worst. The Mimic strategy is volatile and the KNITRO strategy is more stable.

Figure 13 holds the individual task results for human following. Interestingly, the perceived following is more task dependent. The Intent strategy does not strongly follow the human's action for any task but fairs a bit better in the Handover task. The Mimic controller is consistent for all tasks with a high first choice and third choice count. The KNITRO strategy is more robust and is generally perceived to be a good follower of the human across all tasks. This does not necessarily mean the KNITRO strategy outperforms the Mimic, rather it has less variance in the perceived following. This is likely be due to a few factors: 1) the KNITRO strategy listens to an operator's main factors such as palm direction yet does not listen to smaller factors such as position, which ultimately make the difference for evaluators' preference, 2) when addressing this question evaluators subjectively include an ideology of "it does what I want", 3) evaluators could distinguish the subtle differences in how the strategies were following a person and did not prefer the fine tune control the Mimic approach was using. With these trends, the Intent and Mimic strategies have statistical significance for each task, while the Mimic and KNITRO have significance only for the Handover task.

Table VI: Statistical comparison across tasks for following tasks

| Data Breakdown | Intent vs. Mimic | Intent vs KNITRO | Mimic vs KNITRO |
|---|---|---|---|
| Total | **0.0014** | 0.1184 | 0.4430 |
| Usage Only | **0.0015** | 0.7360 | 0.0622 |
| Handover Only | **1.46E-4** | 0.3089 | **0.0453** |
| Relocate Only | **2.41E-4** | 0.2414 | 0.0844 |

3. Overall Preferences

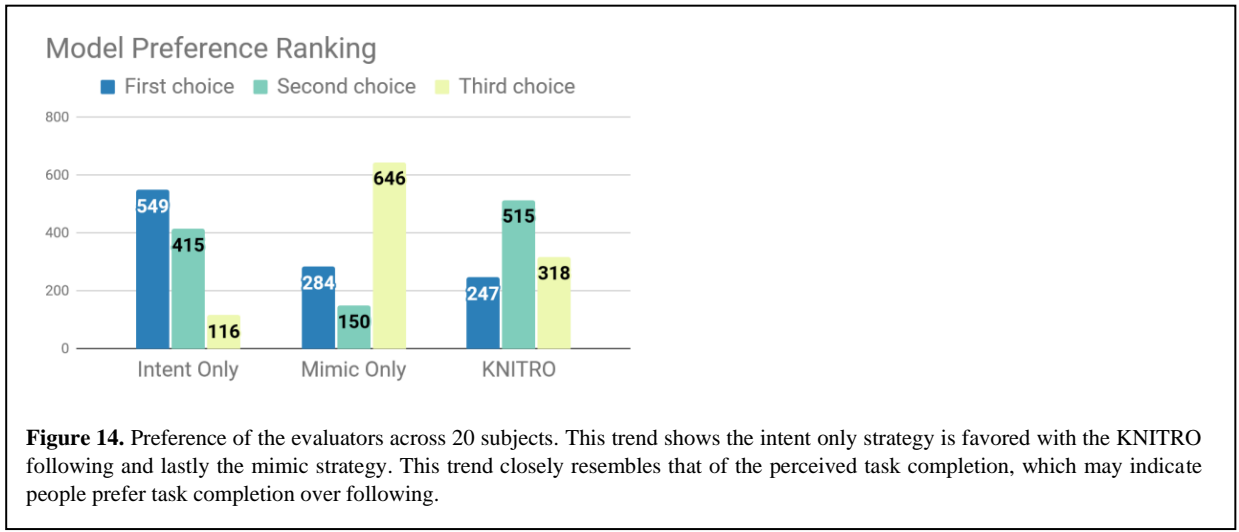

**Figure 14.** Preference of the evaluators across 20 subjects. This trend shows the intent only strategy is favored with the KNITRO following and lastly the mimic strategy. This trend closely resembles that of the perceived task completion, which may indicate people prefer task completion over following.

The ranked choices based on preferences of the bystanders for the total tasks are shown in Figure 14. The Intent strategy is preferred over the other control schemes. The distribution is nearly identical to that of the perceived task completion. The Mimic approach is the least favored but still has high variance with the second most first choices. The KNITRO strategy is in the middle of both other strategies. Table VII shows the Intent and Mimic method have statistical significance. Yet the KNITRO does not have significance with either control scheme.

The preferences based on task are shown in Figure 15. The preferences are consistent across all tasks. The Intent controller is the most favored. The Mimic strategy is the least favored. The KNITRO controller is in between both controllers for the Transfer and Handover tasks and outright beats the Mimic strategy in the Usage task. Table VII shows the statistical significance of the control schemes for the preferences bystanders held. There is always significant difference in preferences between Intent and Mimic strategies, where the Intent is clearly preferred. The KNITRO and Mimic strategy is conditional on the task—significant difference for the Handover task only.

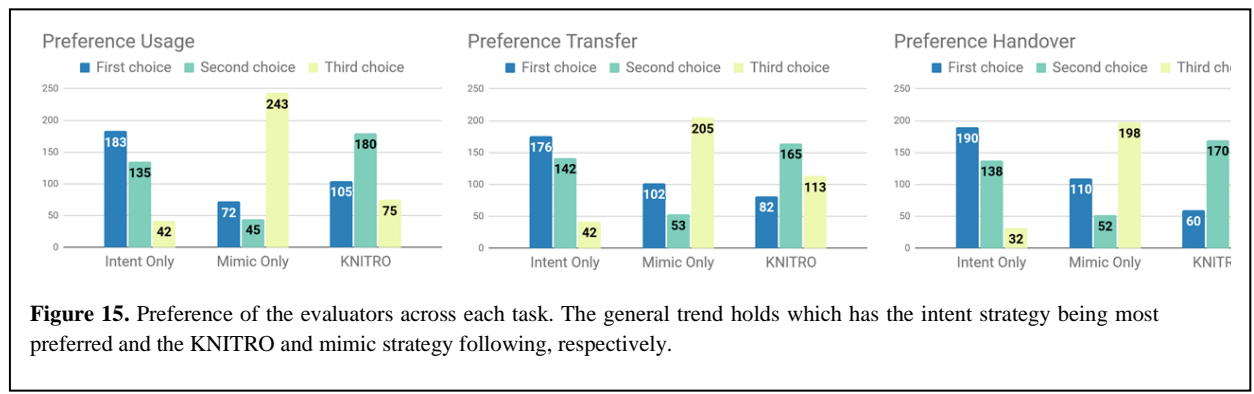

**Figure 15.** Preference of the evaluators across each task. The general trend holds which has the intent strategy being most preferred and the KNITRO and mimic strategy following, respectively.

Table VII: Statistical comparison across tasks for preference

| Data Breakdown | Intent vs. Mimic | Intent vs KNITRO | Mimic vs KNITRO |
|---|---|---|---|
| Total | **4.48E-03** | 7.16E-02 | 2.98E-01 |
| Usage Only | **8.21E-03** | 2.87E-01 | 1.15E-01 |
| Handover Only | **6.74E-04** | 1.72E-01 | **4.18E-02** |
| Relocate Only | **2.84E-03** | 1.53E-01 | 1.20E-01 |

The most interesting trend is observed when comparing Figure 10 and Figure 14. The rankings all bystanders choose are strikingly similar. This suggests bystanders value a successful pose which telegraphs which tasks the robot is accomplishing over a pose which mirrors a human's action. From the bystanders' perspective, it suggests that autonomous robotic assistance is needed so the robot can correctly complete tasks.

### D. Arbitration Guided by Physical Discrepancy

Along with human evaluation, we also analyzed the overall similarity values between different hand structures by calculating the KL divergence values. The values generated in Table VIII are not look up values, rather a validation of our method to show the trend of similarity between hand structures. Since $KL(P\|Q) \neq KL(Q\|P)$, the row indicates P, while the column indicates Q. In a robotics context, agent P will attempt to perform an action based on agent Q's input, where the divergence value associates how similar the models are to one another. The off-diagonal terms do not necessarily need to be equal. For instance, when P= two-finger gripper and Q= three-finger gripper the similarity is (52.66) and vice versa is ($1.18 \times 10^6$). This signifies it is easier for the two-finger gripper model to learn from the three-finger model. Additionally, this table should only be read by comparing columns for a single row. For example, when P= two-finger gripper, where three potential models for Q exist. Of the three potential candidates for Q, we see the three-finger gripper (52.66) is the most similar, while the human hand ($1.79 \times 10^3$) and five-finger gripper ($9.85 \times 10^3$) are orders of magnitude less similar. These differences are likely due to several factors such as the characteristic parameters (size, shape, and number of fingers) as well as grasping configurations (palm orientation, finger contact, and force). The rows can be thought of the hand structure performing the task and the columns are the hand structure providing input on how to complete it. In this instance, the values obtained are only for a single task of transferring a cup, although the trend holds for other tasks. It is interesting to identify how a more complex hand structure is more similar to a relatively simple hand structure unlike the converse case. For instance, the one in our experiment was when P=three finger gripper (second row) and Q=Human hand (first column) with an overall divergence of $1.29 \times 10^3$. Notice how the complex human hand does the best job in relating to the three-finger gripper despite nearly being the same robot as the two-finger gripper. When observing the last row where P= five-finger gripper, the divergence values show the complex human hand (14.14) is most similar to the five-finger robot hand, while the three-finger gripper ($1.14 \times 10^5$) and the two finger gripper ($9.0 \times 10^5$) are more divergent.

**Table VIII.** Example of divergence values for different hand structures

| P: row, Q: column $KL(P\|Q)$ | Human Hand | Two-Finger Gripper | Three-Finger Gripper | Five-Finger Gripper |
|---|---|---|---|---|
| Human Hand |  | $1.18 \times 10^6$ | $9.6 \times 10^5$ | 122.49 |
| Two-Finger Gripper | $1.79 \times 10^3$ |  | 52.66 | $9.85 \times 10^3$ |
| Three-Finger Gripper | $1.29 \times 10^3$ | $1.18 \times 10^6$ |  | $3.26 \times 10^3$ |
| Five-Finger Gripper | 14.14 | $1.14 \times 10^6$ | $9.0 \times 10^5$ |  |

Discussion:

A. Operator Analysis

The operators prefer assistance over direct telemanipulation. Not only the objective measures of time and success demonstrate the efficacy of intent based shared control, but the subjective results as well. The KNITRO approach is slightly favored over the Intent Only controller on the subjective evaluation. The Intent strategy inherently follows an operator due to the inference models, yet it may feel disconnected at times if an operator encounters an unexpected motion (a rotation or sudden jump in position). This is where the KNITRO controller may thrive since the robot will not stray from the operators' important input motion, such as palm direction, yet adjust other variables such position or finger control. It makes intuitive sense that the more autonomy we inject into the system the faster the system can achieve the task. Yet, too much autonomy can result in a loss of control as it ignores the explicit commands of the operator by producing unexpected movements, which leads to an overall distrust of the system. These potentially unexpected movements cause operators to overreact in their own adaption, thus changing the inputs sent to the robot, and finally resulting in the robot failing the task altogether. Therefore, we recommend the Intent Only strategy should be reserved for control where an operator does not provide direct inputs, rather more high-level or partial control such as eye gaze only control (Li and Zhang (2017); Li et al. (2017)), or in situations where the operator has a lack of understanding of the dynamics of the robot (Reddy et al. (2018a)). For these reasons, we believe the KNITRO strategy is most appropriate moving forward when an operator can give more explicit commands (instead of position only control, they also give orientation, force, etc.) and understand the dynamics better since it provided the strongest preference, felt the most natural, yielded the best success rate albeit at the sacrifice of some speed (compared to the Intent Only controller).

In this paper we used a common object with simple interaction rules to analyze the methods, where we see the clear expected trend. These trends we observe are possibly more significant depending on the irregular handling of a shape, in which the task difficulty will be higher and the operator would rely more heavily on the assistance of the robot. Therefore, there is more room for improvement through operator personalization of the penalty terms, $\lambda_i$, $\gamma$. Although we present a single approach to impact the arbitration through physical discrepancy, we acknowledge there are many other factors which are important to consider for a truer representation of the system (Young et al. (2019)). For instance, if a user wants to focus only on position control and allow the robot to adjust the orientation independently, they should be allowed. Users may have different preferences for what is considered natural, or an intuitive response to the robot. This type of constraint is already possible within the framework of the control system proposed by empirically tuning the penalty terms. Fixed values are not the only option as dynamic values are viable and encouraged, in that it is possible to design a function and adjust the arbitration terms based on historic performance/preference of the operator. In fact, task difficulty may demand it based on the results presented. During the trials, participants often noted during the transfer task (easiest of the three), the assistance was not as apparent; however, they would later remark for the harder tasks they felt the robot adjusting greatly (in a positive view) to help them complete the task. This would lead to credence that the penalty terms should be a dynamic function with an input regarding task difficulty.

### B.  Strategy Legibility and Task Completion

Although others attempt to disambiguate and use separate models depending on the disambiguated task, we do not believe it is possible to always perform these types of methods for all situations. In (Javdani et al. (2015)), they attempt to disambiguate the target object out of a set of objects with discrete states for approaching tasks, however, this discretization does not exist in grasping. The strategies presented in our paper are used to deal with ambiguity from human operators to robot systems since this ambiguity is inherent and present in a majority of practical scenarios. Thus, the burden of dealing with the uncertainty from the human input and the robot environment should be placed on the robot instead of the human. The robot implicitly must achieve two criteria: 1) complete the inferred task from the person, and 2) display legibility to the operator so that she/he knows what task it is accomplishing. This naturally forms a shared control problem. Our approach does arbitration by considering physical discrepancy to achieve the tradeoff between these two criteria. The KNITRO strategy attempts to achieve both goals to determine what people prefer from robotic assistant systems. The Mimic method should theoretically carry the highest level of legibility to the operator as it strictly/explicitly follows the person, however, this makes the Mimic method extremely volatile to task success as it requires a well experienced operator to ensure the task is successful despite not fully understanding the robot environment. This is where the other two approaches may outperform the Mimic strategy since the level of assistance is increased.

The Intent strategy is a form of the KNITRO strategy in the sense that an inherent level of legibility exists with the strategy. However, the level of legibility is limited as it is implicit, rather than the more explicit approach of the KNITRO or Mimic methods. It is possible the Intent and KNITRO approaches do show similar responses when the desired human approach aligns with the rules which the robot model knows and attempts to follow. The KNITRO strategy is designed to objectively make a compromise between the tradeoffs of both the Intent and Mimic strategies to reduce the mental workload of the user. Additionally, the KNITRO formulation is generalizable in the sense where now a formal framework has been introduced, other potential methods exist in determining the weighting scheme to create desired behaviors.

Hypotheses

The first hypothesis was to determine if assistance improves objective outcomes. The results indicate this is true albeit task difficulty does appear to exasperate the degree. The KNITRO improves the success rate over the Mimic strategy more than the Intent Only controller was able to improve over the Mimic strategy. This is likely due to a strong counter reaction by the operator as they perceive a loss of control of the robot as it gains this extra autonomy. When considering the completion time, the Intent Only controller shines. Adding the extra autonomy allows the operators to complete the task much quicker.

The second hypothesis was to compare the preference of the assistance modes and determine if the KNITRO is preferrable to the Intent Only controller. The results are inconclusive when considering both the operators and bystanders. The operators have a slight preference to the KNITRO strategy, while the bystanders prefer the Intent Only. The perception of the operators should be weighted higher, thus proving the hypothesis. However, the effects of a bystander being distracted by less legible motion needs to be further investigated in a new study before this claim can be considered true. However, this difference does support the third hypothesis. This hypothesis was to determine if there was a preference difference between operators and bystanders. The results indicate this is true. Operators prefer the KNITRO as they have a say over how the robot moves. The actions they take may be easier for the operator to move the robot but lack legibility of the task. The bystanders prefer the Intent strategy as its final pose indicates to the bystanders which task is being completed. Moving forward, the discrepancy between the operators and bystanders is imperative to account for when designing future control schemes.

Funding:

This material is based on work supported by the US NSF under grant 1652454. Any opinions, findings, and conclusions or recommendations expressed in this material are those of the authors and do not necessarily reflect those of the National Science Foundation.